\documentclass{article}
 
\usepackage{arxiv}
\usepackage[utf8]{inputenc} % allow utf-8 input
\usepackage[T1]{fontenc}    % use 8-bit T1 fonts
\usepackage{times}
\usepackage{epsfig}
\usepackage{graphicx}
\usepackage{amsmath}
\usepackage{amssymb}
\usepackage{xcolor}
\usepackage{bm}
\usepackage{float}
\usepackage{url}            % simple URL typesetting
\usepackage{hyperref}
\usepackage[linesnumbered,algoruled,boxed,lined]{algorithm2e}
\SetKwInOut{Parameter}{Parameters}

\usepackage{subfigure}
\usepackage{booktabs}       % professional-quality tables
\usepackage{amsfonts}       % blackboard math symbols
\usepackage{nicefrac}       % compact symbols for 1/2, etc.
\usepackage{microtype}      % microtypography

\usepackage{etoolbox} % for "\patchcmd" macro
\makeatletter
\patchcmd{\ps@headings}{\rlap{\thepage}}{}{}{}
\patchcmd{\ps@headings}{\llap{\thepage}}{}{}{}
\makeatother
\begin{document}

%%%%%%%%% TITLE
\title{On the Feasibility and Generality of Patch-based Adversarial Attacks on Semantic Segmentation Problems}

\author{Soma Kontár, András Horváth\\
Peter Pazmany Catholic University Faculty of Information Technology and Bionics\\
Budapest, Práter u. 50/A, 1083
}

\maketitle

\begin{abstract}
Deep neural networks were applied with success in a myriad of applications, but in safety critical use cases adversarial attacks still pose a significant threat.
These attacks were demonstrated on various classification and detection tasks and are usually considered general in a sense that arbitrary network outputs can be generated by them.

In this paper we will demonstrate through simple case studies both in simulation and in real-life, that patch based attacks can be utilised to alter the output of segmentation networks.
Through a few examples and the investigation of network complexity, we will also demonstrate that the number of possible output maps which can be generated via patch-based attacks of a given size is typically smaller than the area they effect or areas which should be attacked in case of practical applications.

We will prove that based on these results most patch-based attacks cannot be general in practice, namely they can not generate arbitrary output maps or if they could, they are spatially limited and this limit is significantly smaller than the receptive field of the patches.
\end{abstract}

\section{Introduction}
With the application of deep neural networks becoming mainstream in our everyday lives, questions and concerns about the robustness and reliability of these networks are also becoming ever more important.
Adversarial attacks targeting the vulnerabilities of neural networks were investigated heavily in the past years.
These attacks are considered general in the sense that with the proper optimization techniques arbitrary outputs can be generated by them, regardless of the input image, which means that these attacks pose a significant threat in practical applications.

Adversarial attacks were first introduced in \cite{szegedy2013intriguing} and they have revealed an important aspect of deep neural networks: although they generalise well and work properly not just on the typical input set, but also on similar inputs, they can be exploited by malevolent attackers, since inputs are high dimensional and one can easily generate non-real life samples, which fall extremely far from both human judgement and the expected outcome.

In the following years, the possibilities of exploiting adversarial attacks were investigated elaborately, building on the original authors' findings \cite{goodfellow2014explaining}, \cite{rozsa2016adversarial}, \cite{dong2018boosting}, \cite{moosavi2016deepfool}, devising new attack strategies improving the robustness of the generated attacks \cite{brown2017adversarial}, \cite{athalye2017synthesizing}, enabling black-box attacks in which case the gradients of the network are not necessary \cite{eykholt2017robust}, \cite{alzantot2018genattack}, \cite{papernot2017practical}.
Later, adversarial attacks were also presented on more complex tasks than classification, like detection and localisation problems \cite{thys2019fooling}, and on various network architectures (e.g.: Faster-RCNN \cite{chen2018shapeshifter}). 

The first attacks in classification and detection problems applied minor, low-intensity perturbations over the whole image, just like in \cite{goodfellow2014explaining} for classification. It was later demonstrated that low-intensity attacks are not at all robust in the wild \cite{lu2017no}.
In practice this special additive noise is usually altered by perspective distortion and additive noise in the environment (e.g.: illumination changes) and also not life-like, since in practice the attacker needs to have access to the image processing system to modify all elements of the input, instead of modifying a real-world object. 

Although real life and robust attacks were demonstrated for classification and detection problems, in case of segmentation problems, only low intensity attacks were investigated \cite{xie2017adversarial}, \cite{metzen2017universal}. Segmentation problems are also more complex and their output depends on fine details of the input samples. In case of classification one expects that the output class should not depend at all from the pose of the investigated object and in case of detection problems only small changes should appear. If the object rotates slightly or changes its pose (e.g.: a person moves his arm) output classes should remain exactly the same and bounding-boxes should change slightly, meanwhile segmentation masks might change rather significantly. Based on this one could expect that segmentation networks are more robust toward real-life adversarial attacks.

It was demonstrated in \cite{xie2017adversarial} and \cite{metzen2017universal} that networks trained for semantic segmentation problems can also be attacked with low-intensity noise and the authors could generate arbitrary output maps with the proper additive noise. Although it was never proven that these methods can generate arbitrary output maps, the authors have demonstrated that highly uncorrelated and randomly selected output maps could be achieved, which gave rise to the general belief in the scientific community that these low-intensity approaches can result in arbitrary output maps.

In \cite{nakka2020indirect} it is shown that state-of-the-art semantic segmentation networks are vulnerable to some indirect local adversarial attacks -- in the attack scenario a patch is placed in the environment, creating "dead zones" for a particular class of objects. While this does show that some networks are vulnerable to patch-based adversarial attacks, the authors found that models with a bigger field of view are more sensitive to these kinds of attacks. In contrast, our method is closer to a real-life scenario, since we are only modifying the object the attack is targeting, thus not needing access to the environment we are performing our attack in.
%We should refer to this at the complexity analysis as well, since the receptive field is even larger in this case

The authors of this paper are not aware of any successful direct patch based attacks on semantic segmentation problems, which emphasises the difficulty of generating such samples.

In this paper we will demonstrate that patch based attacks are  feasible on semantic segmentation problems. After demonstrating their feasibility we will analyze their generality. The number of possible output maps using a patch of limited size is typically fairly small, which means that these attacks -- contrary to general belief -- can not be used to generate arbitrary outputs, driving the conclusion that they cannot be general.

We also have to admit that the non-generality of patch based attacks on segmentation problems does not mean at all that they can not be applied in practice. This only means that arbitrary outputs can not be generated by them, but the question of which outputs can and can not be generated by an adversarial patch remains open which the authors plan to investigate in their future work.

\section{Adversarial Attacks}\label{AdvAttacks}

The term \textit{adversarial example} was coined by \cite{szegedy2013intriguing}, where attacks on neural networks trained for image classification were generated via a very low intensity, specially formed additive noise, completely imperceptible to the human observer. The method used to generate these so-called adversarial examples was to maximize the networks response to a certain class by altering the input image. 

The first attacks \cite{goodfellow2014explaining} were implemented by calculating the sign of the elements of the gradient of the cost function ($J$) with respect to the input ($\bm{x}$) and expected output ($y$), multiplied by a constant to scale the intensity of the noise (formally $\epsilon\text{sign}\nabla_xJ(\bm{\theta}, \bm{x}, y))$, where $\bm{\theta}$ is the parameter vector of the model). This allows for much faster generation of attacks. This method is called the Fast Gradient Sign Method (FGSM).

An extension to FGSM by \cite{rozsa2016adversarial} was to use not only the sign of the raw gradient of the loss, but rather a scaled version of the gradient's raw value. This method is usually referred to as the Fast Gradient Value (FGV) method.

Another extension to the iterative version of FGSM by \cite{dong2018boosting} was to incorporate momentum into the equation, theorizing that similarly to 'regular' optimization during training, it would help avoiding poor local minima and other non-convex patterns in the objective function's landscape.

\cite{moosavi2016deepfool} builds on the assumption that the robustness of a binary classifier $f$ at point $\bm{x}_0$, is equal to the distance of $\bm{x}_0$ from the separating hyperplane $\Delta(\bm{x}_0;f)$. Therefore the necessary smallest perturbation to change the sign of the output of $f$ corresponds to the orthogonal projection of $\bm{x}_0$ onto the separating hyperplane. They solve this in a closed-form formula, and apply these small perturbations to the image in an iterative manner until the decision of the classifier is changed. Later they extended it to multiclass classification problems as well.

Though these approaches were extremely important from a theoretical point of view and the generation methods are general, they pose no significant threat to practical applications of neural network, since they limit the amount of applied noise. The smallest perturbations beside the engineered additive noise, e.g.: perspective or illumination changes, lens distortion could completely upend the desired results. Hence, the application of these attacks in real life is unfeasible\cite{lu2017no}.

%\color{red}TODO\color{black}one-step attack is easy to transfer but also easy to defend” \cite{kurakin2016adversarial}

In \cite{brown2017adversarial}, \cite{athalye2017synthesizing} robust and real-world attacks were presented against various classification networks. These methods create an adversarial patch, where instead of the global, but low-intensity approaches, distortions appear in a region with limited area, but intensity values are not bounded\footnote{apart from the global bounds of image values}. 
Successful attacks with adversarial patches were also demonstrated using black and white patches only \cite{eykholt2017robust}, where not the intensities of the patch, but the locations and sizes of the stickers are optimized. These attacks, where the gradients of the networks are not necessarily used during optimization open space towards black-box attacks \cite{alzantot2018genattack}, \cite{papernot2017practical}, where the attacker needs access only to the final responses, confidence values to generate attacks using evolutionary algorithms.
Later these approaches were presented on detection and localization problems as well \cite{thys2019fooling} using various network architectures (e.g.: Faster-RCNN \cite{chen2018shapeshifter} ). 

A general overview of adversarial attacks, containing a more detailed description of most of the previously mentioned methods can be found in the following survey paper \cite{akhtar2018threat}.
The resilience of segmentation networks against adversarial attacks was investigated heavily in the past years \cite{xie2017adversarial}, \cite{metzen2017universal}, \cite{arnab2018robustness}, \cite{al2021class}. But only global, low-intensity attacks were examined. The authors are not aware of any publication demonstrating patch based attacks on semantic segmentation.

\section{Patch-based Segmentation Attacks on a Simple Dataset}\label{ClevR}

Since we were not able to find a simple segmentation dataset (like MNIST\cite{lecun1998mnist} for classification), where objects with various shapes can be found we have created a simple dataset based on CLEVR\cite{johnson2017clevr}. The original dataset did not contain segmentation masks but was used for visual question answering. We have modified the generator script and generated masks for semantic, amodal and instance segmentation.

The dataset contains 25200 colored images (of size $320 \times 240$) of simple objects along with their instance masks, amodal masks and pairwise occlusions and three-dimensional coordinates for each object.
This results a simple dataset for various tasks, ranging from three-dimensional reconstruction and instance segmentation to amodal segmentation.
The dataset contains objects of simple shapes, but also contains shadows, reflections and different illuminations, which make it relevant for the evaluation of segmentation algorithms.

The dataset along with the script and all our training codes belonging to later chapter is available at the following \href{https://drive.google.com/drive/folders/1UzozXfsuOb-IpUGF-9YrfIVN1WrrAWsw}{repository} to help reproducibility and the detailed investigation of the applied parameters.

We have selected the U-net architecture\cite{ronneberger2015u} to be trained on our simple CLEVR inspired dataset. Our aim was to demonstrate adversarial attacks with architectures where classification and segmentation are handled in the same layers and not by different heads, as in case of Mask-RCNN\cite{he2017mask}, where the classifier head might be fooled by the attack, meanwhile the network provides a same or similar instance mask.
We have used a U-NET like structure containing convolution blocks with 8, 16, 32, 64 channels,
downscaling was accomplished by strided convolution, while upscaling was implemented by transposed convolution.
23400 images were selected for training and 1800 images were left for validation. All the validation scenes were generated independently from the training scenes.
Our selected task was semantic segmentation, where a four-channeled output image had to be generated by the network representing the probabilities, that a pixel belongs to a cube, sphere, cylinder or to the background. 

\begin{figure}[!htp]
\centering
\subfigure{\includegraphics[width=0.245\textwidth]{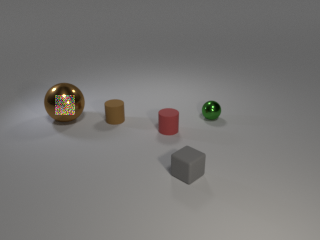}}
\subfigure{\includegraphics[width=0.245\textwidth]{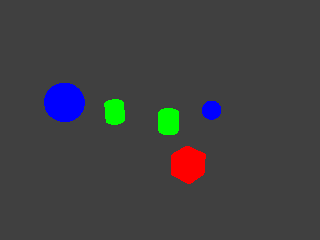}}
\subfigure{\includegraphics[width=0.245\textwidth]{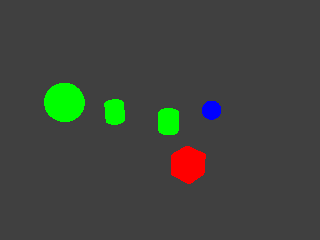}}
\subfigure{\includegraphics[width=0.245\textwidth]{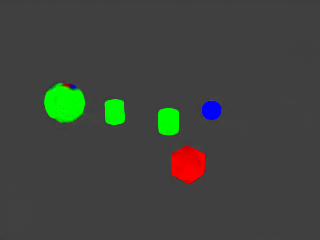}}\\
\subfigure{\includegraphics[width=0.245\textwidth]{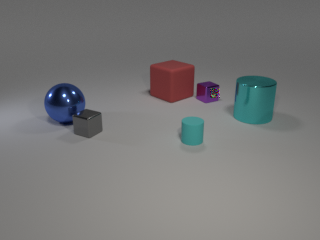}}
\subfigure{\includegraphics[width=0.245\textwidth]{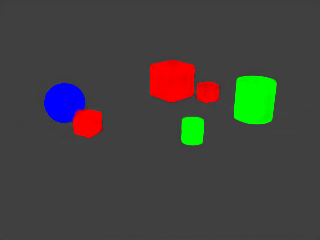}}
\subfigure{\includegraphics[width=0.245\textwidth]{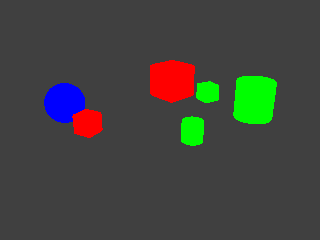}}
\subfigure{\includegraphics[width=0.245\textwidth]{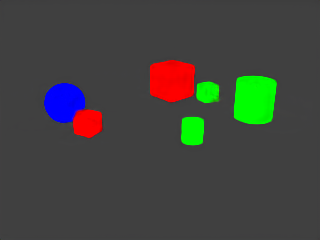}}\\
\subfigure{\includegraphics[width=0.245\textwidth]{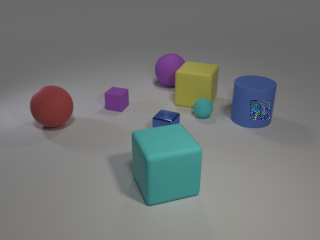}}
\subfigure{\includegraphics[width=0.245\textwidth]{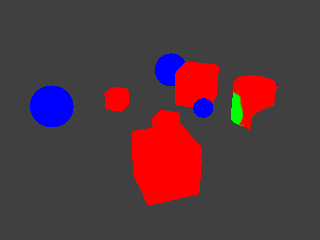}}
\subfigure{\includegraphics[width=0.245\textwidth]{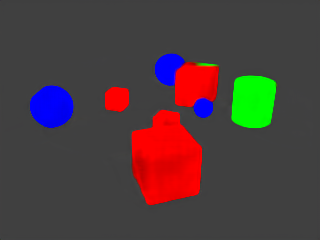}}
\subfigure{\includegraphics[width=0.245\textwidth]{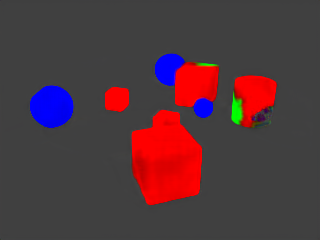}}
\caption{This figure depicts segmentation attacks on the CLEVER dataset. The first column contains the attacked image, after the patch was optimized for this sample for 5000 iterations. The second column displays the expected outputs after the attacks, meanwhile the third and fourth columns show the network output before and after the attack. The first two rows contain samples where object classes were switched and the last row contains a sample, where the mask was drawn by hand. Please note that part of the output mask was deleted, to demonstrate that objects can not only be turned to other classes, but they can partially or completely disappear. We have to note that there is no theoretical difference between modifying an output pixel to belong to a desired class or to none of them.}
\label{FigClevrAttacks}
\end{figure}

We have selected 100 samples from our test set randomly and trained adversarial patches using the method published in \cite{dong2018boosting}. We have tried scripted attacks where we have changed the expected output class of a selected object (in this case we did not change the shape of the mask), but even in this case the network has no advantage from the previously learned shapes or from the fact that objects in our database has consistent shapes, since here the aim was to segment spheres with shape of a cube or cylinder.
But to prove that the network can create arbitrary shapes and segments, we have also created expected masks by hand and tested our approach on them. We have opted for the previously mentioned, scripted method in larger case experiments because class switching could be easily implemented by scripts.
Sample attacks with the expected masks and network outputs before and after the attacks can be found on Fig. \ref{FigClevrAttacks}

As it can be seen from the previous examples, patch based attacks were possible on this simple segmentation task. The outcome of the network was close to the expected mask in almost all examined cases and even though they were not perfectly reconstructed every time, altogether $95\%$ of the output pixels were modified as expected.

\subsection{Real-life Images of Simple Shapes}

Once we have created successful attacks in simulation we gathered 100 real life samples (10 different setups from 10 different views) and tested our approach on them. We have applied our method without fine-tuning or further optimisation on real samples or without the application of any domain adaptation methods \cite{ganin2016domain}, \cite{tzeng2018splat}. Our network worked on an acceptable level on real-life samples. Although segmentation was noisy and many small objects have appeared in the background, segmentation of real objects was correct regarding shapes and classes, which are the most important to investigate possible attacks. 

We have selected 10 images and changed their outputs by hand: we have repainted one of the objects on the segmented output map and used the same method as in case of the simulated data on these specific images to change the output of the network.

To demonstrate the robustness of the patches in this experiment we have followed the generation of the patches introduced in \cite{brown2017adversarial}. For training we have created new simulated data, with added variance on view-angles, scales (camera distances from the objects) and lighting conditions, but all images contained the objects in the same constellation. Later we have used these images an tried to generate one patch which works well on the selected object regardless of the previously listed variances. We have also added small random noise to the intensity values of the patch and changed the position of the patch on the image slightly to avoid the generation of a low-intensity attack. Additionally to random disposition of the patch we have applied average pooling on it with a kernel of $3\times3$ and stride one, this way we have optimized the average of neighbouring patch pixels in each kernel, instead of directly optimizing each pixel and we got more consistent and better results.

\begin{figure}[!htp]
\centering
\subfigure{\includegraphics[width=0.245\textwidth]{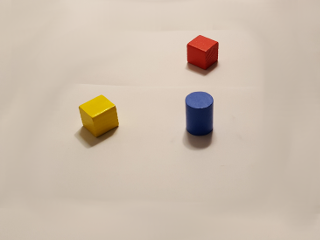}}
\subfigure{\includegraphics[width=0.245\textwidth]{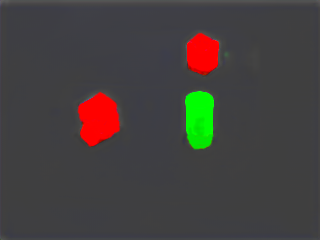}}
\subfigure{\includegraphics[width=0.245\textwidth]{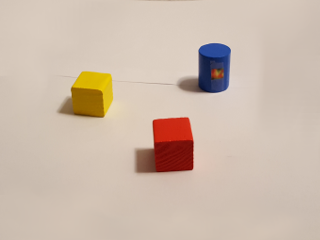}}
\subfigure{\includegraphics[width=0.245\textwidth]{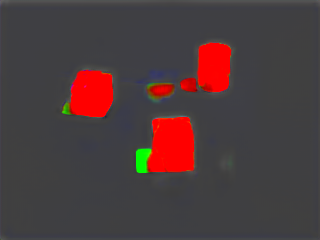}}
\caption{This figure presents the effect of a printed patch in real life in segmentation. as it can be seen we have managed to train an adversarial patch in simulation which was able to turn a cylinder into a cube in a real life segmentation problem.} \label{PrintedATtacks}
\end{figure}

In all the previous cases patches were created and added to the image in simulation, in this case we have optimized a patch to turn a cylinder to a cube on our simulated dataset, but printed it out and tested it on real life samples. A randomly selected example from our samples can be seen on Fig. \ref{PrintedATtacks}. As it can be seen the blue cylinder was segmented correctly without the patch, but when the patch was attached to it its segmentation turned red signifying its pixels belong to a cube.
We have to note that a more thorough study on more samples and with more complex examples is needed to understand how patch based attack can be efficiently generated in real-life, but our experiments demonstrate that robust adversarial patches, which are applicable under multiple views and conditions, are feasible.

\section{Patch-based Segmentation Attacks on Cityscapes}\label{RealAttack}

For a more complex case study we have chosen the Cityscapes dataset\cite{Cordts2016Cityscapes}  focusing on the semantic segmentation task of the dataset. Our models of choice were the Deeplab V3 \cite{chen2017rethinking} and MobileNet V3 architectures\cite{hu2018squeeze}, in both of which a significant feature is that the mask prediction and the classification are calculated jointly, unlike in case of Mask-R-CNN\cite{he2017mask}.

In our experiments, we used a ResNet-18 backbone for the Deeplab V3 architecture and MobileNet V3 Large network, with 128 filters in the segmentation head. For both models, we utilized openly available pre-trained models, the Deeplab V3 model is available \href{https://github.com/fregu856/deeplabv3} {on Github}, while the MobileNet V3 model is available via \href{https://pypi.org/project/fastseg/} {PyPi}.

\begin{figure}[ht]
\begin{center}
%\captionsetup{justification=centering}
\subfigure{    \includegraphics[width=0.23\textwidth]{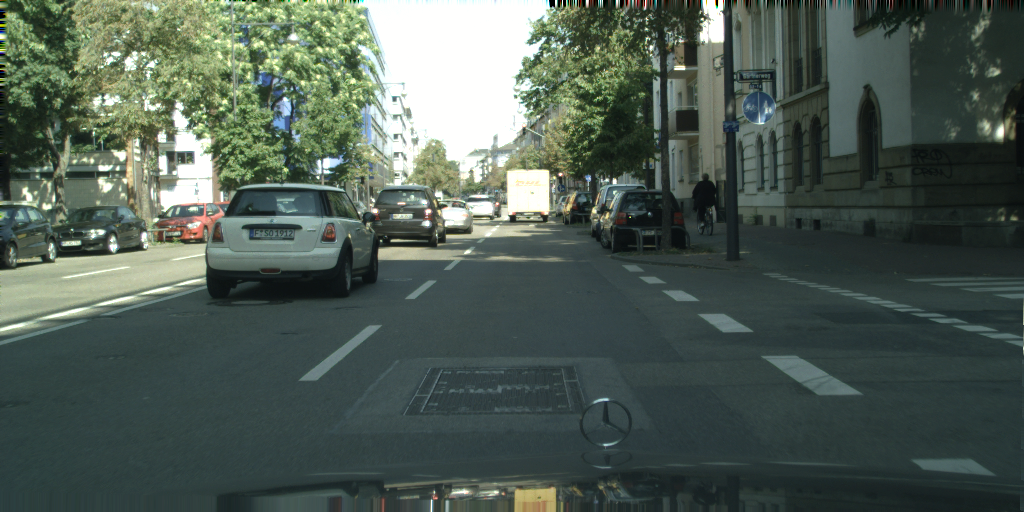}}
\subfigure{     \includegraphics[width=0.23\textwidth]{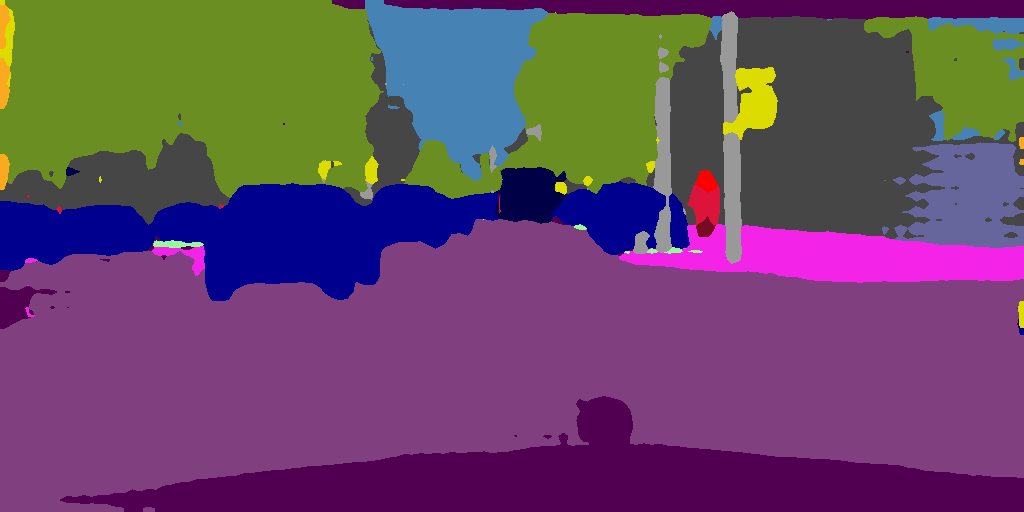}}
\subfigure{    \includegraphics[width=0.23\textwidth]{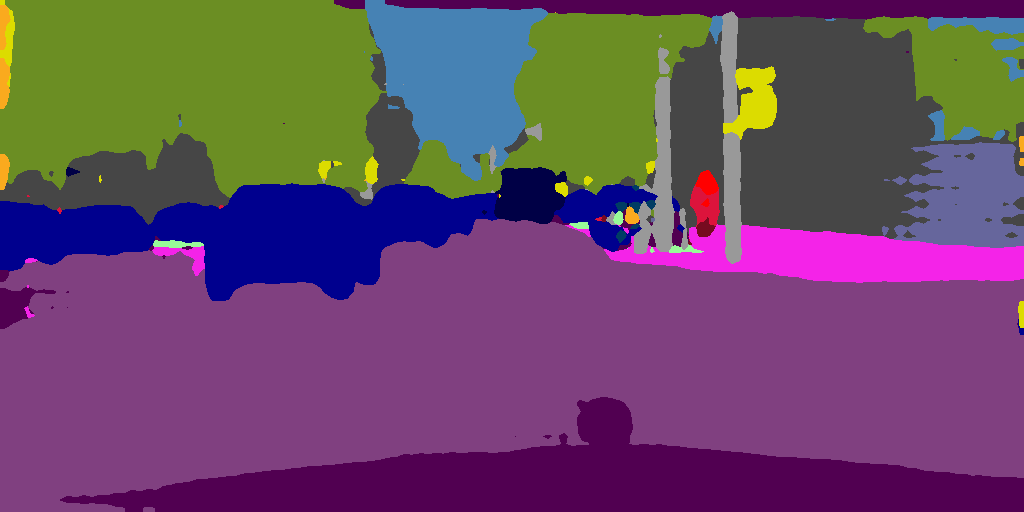}}
\subfigure{     \includegraphics[width=0.23\textwidth]{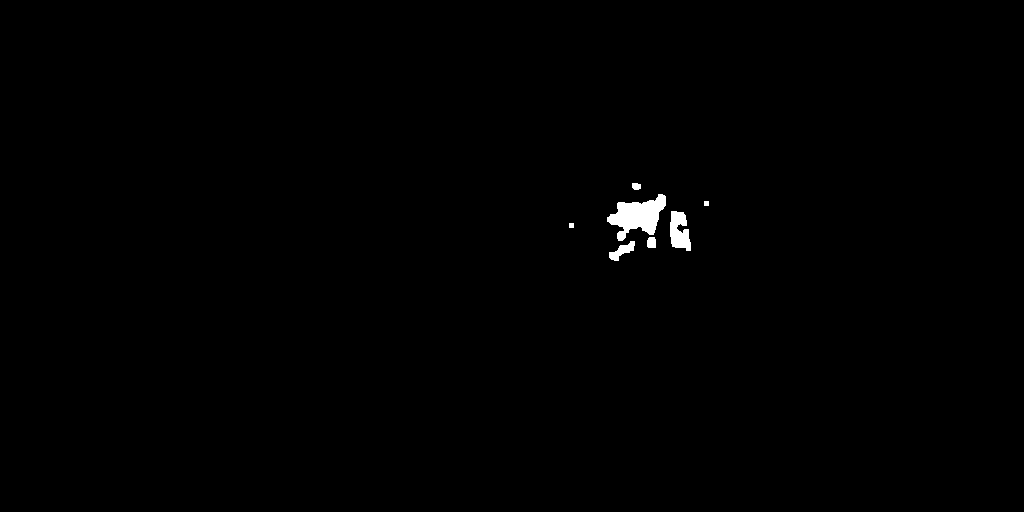}}\\
\subfigure{    \includegraphics[width=0.23\textwidth]{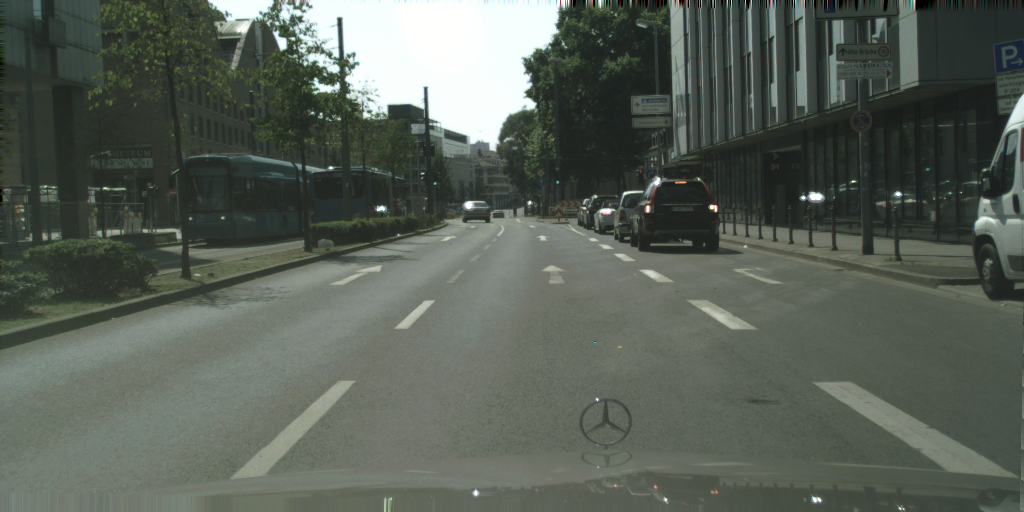}}
\subfigure{     \includegraphics[width=0.23\textwidth]{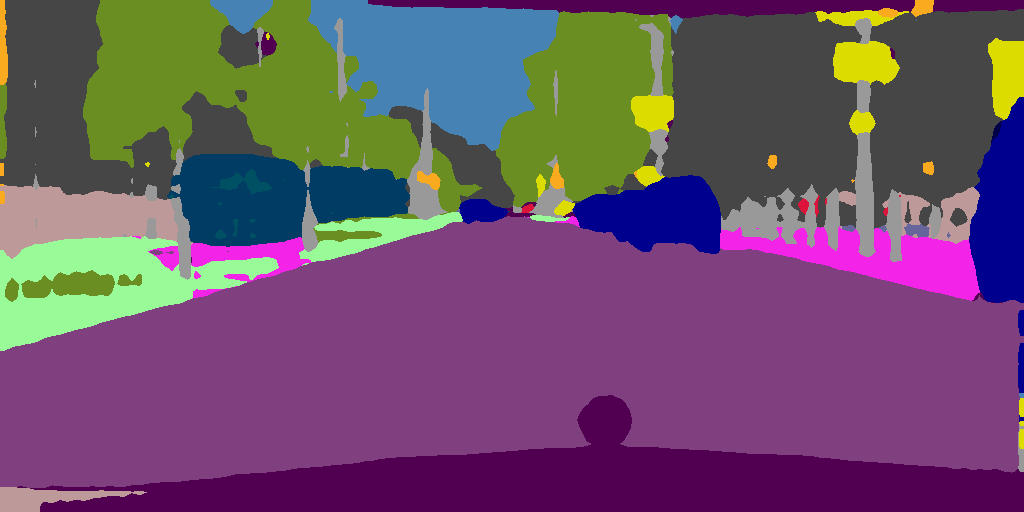}}
\subfigure{    \includegraphics[width=0.23\textwidth]{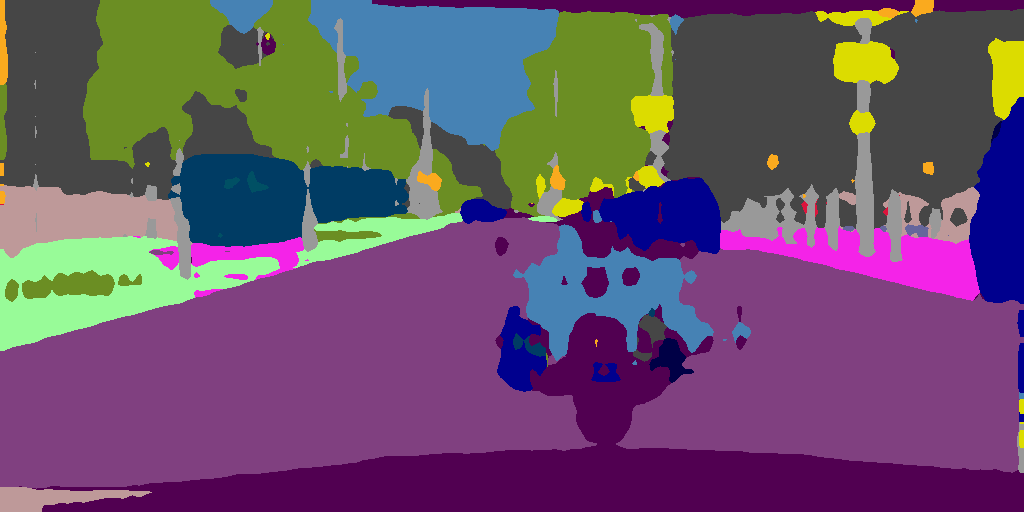}}
\subfigure{     \includegraphics[width=0.23\textwidth]{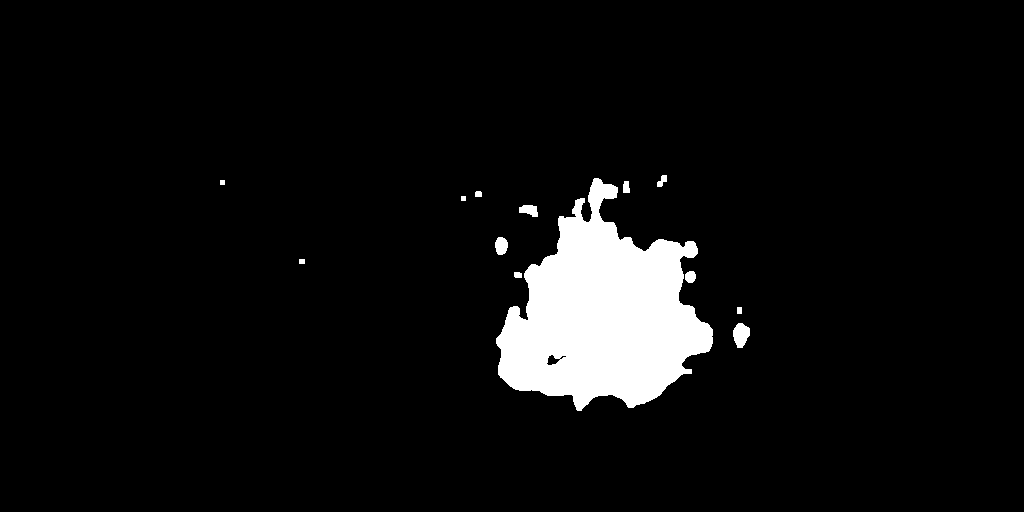}}\\
\subfigure{    \includegraphics[width=0.15\textwidth]{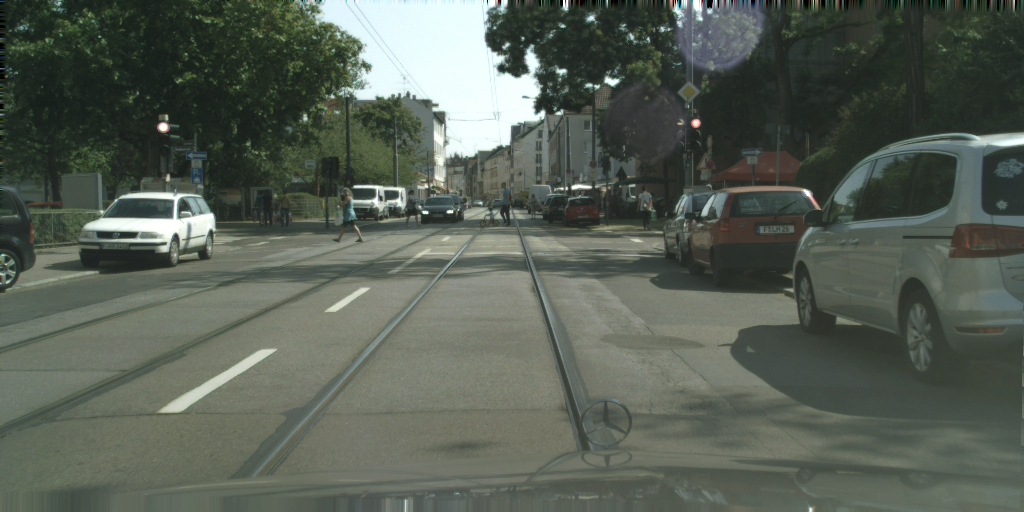}}
\subfigure{     \includegraphics[width=0.15\textwidth]{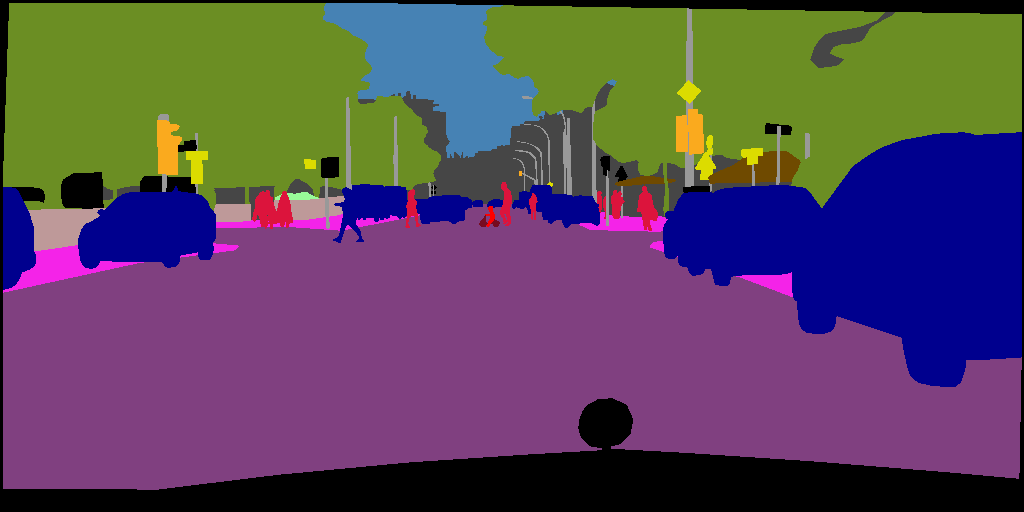}} 
\subfigure{    \includegraphics[width=0.15\textwidth]{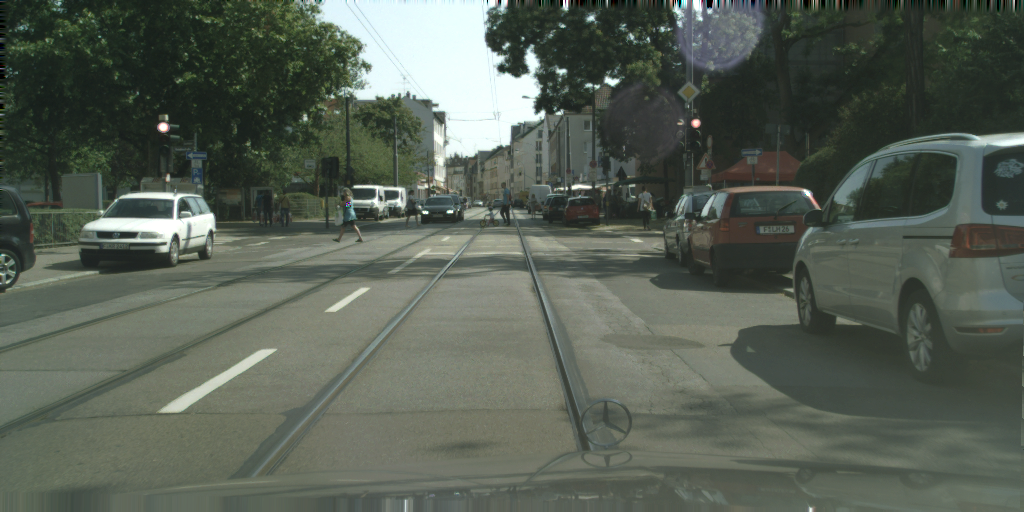}}
\subfigure{     \includegraphics[width=0.15\textwidth]{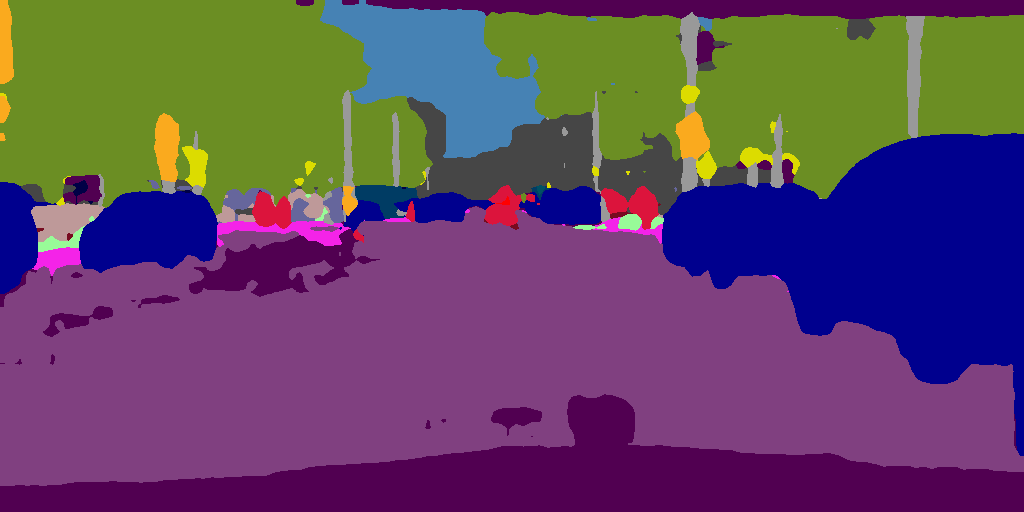}} 
\subfigure{    \includegraphics[width=0.15\textwidth]{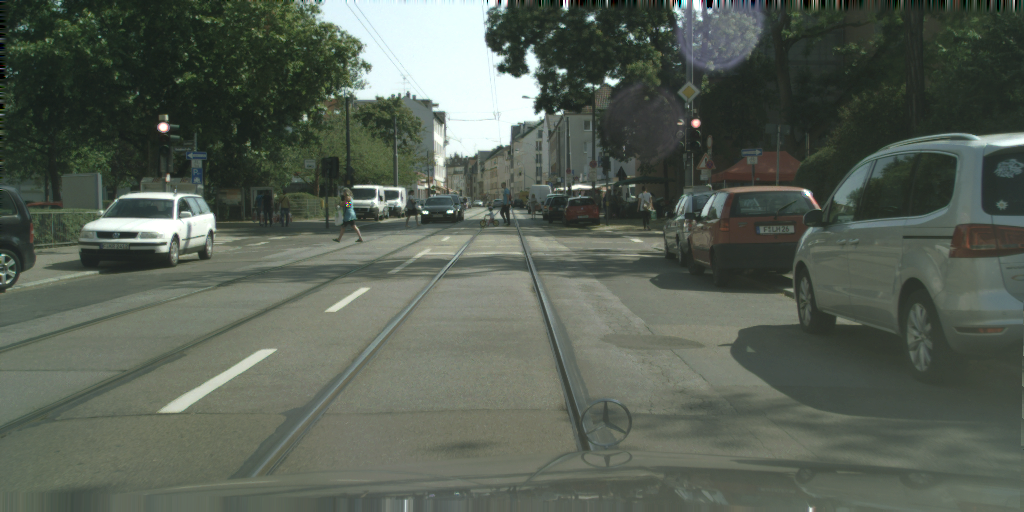}}
\subfigure{     \includegraphics[width=0.15\textwidth]{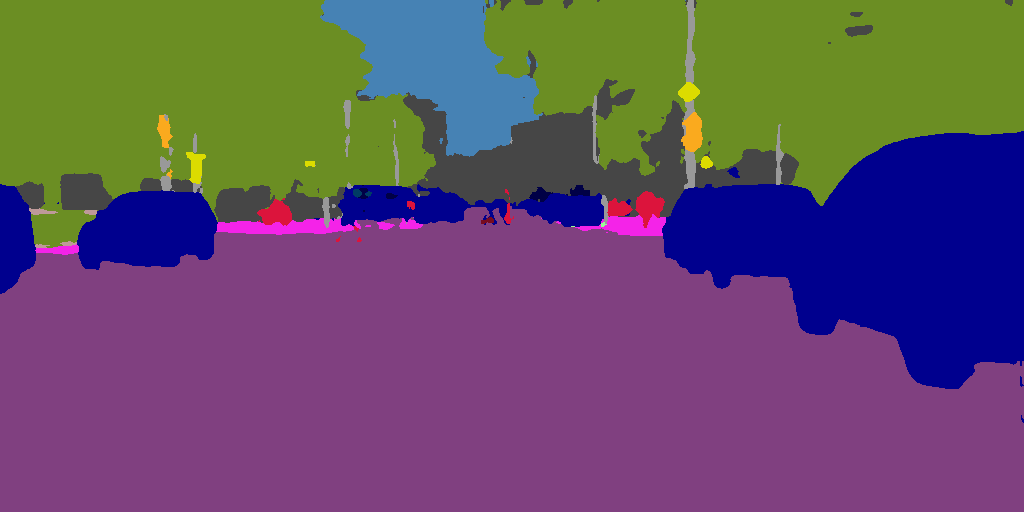}}
\end{center} 
\caption{Example attacks on the Cityscapes dataset. The top row depicts details of an attack on the DeeplabV3 architecture where the pixels of the car should be turned into arbitrary other classes. The first column contains the input image with a 2x2 patch on the left lamp of the parking car and the second column display original output of the trained network without the adversarial patch. The next image displays the output of the network for the patched image and the effect of the patch, those pixels are marked by white where the output class (after argmax) of a pixel have changed are marked on the last image. As it can be seen from this figure the effect of 2x2 patch is fairly large and it has changed the output class of 7461 pixels.The second row contains a similar example with the Mobilenet V3 architecture where an imaginary wall should be placed in the middle of the road. The last row depicts an other attack scenario where a pedestrian is completely removed from the network's output. The row's layout is the following left to right, by rows: original input, target mask, Deeplab V3 patched input, Deeplab V3 attack result, MobileNet V3 patched input and finally MobileNet V3 attack result. This not only demonstrates the vulnerability of neural networks that are used in mission-critical applications (ie. self-driving), but also signifies that adversarial attacks pose a significant threat in security applications as well. }
\label{fig:alg.illustration}
\end{figure}

For the generation of the adversarial patch we have followed the method described in \cite{brown2017adversarial}, in which the adversarial patch, which instead of modifying every pixel of the image with an additive noise completely replaces a small part of the image, is trained in a white-box setting by minimizing a loss between a (usually hand-crafted) target
and the output of the network by only altering the values in the patch arbitrarily\footnote{The values are of course bounded within the regular image values before preprocessing, eg. [0, 255].}. In our experiments, we have first selected a class and for the sake of reproducibility we used an algorithm to find the largest inscribable rectangle for a given binary object mask of this class, which can easily be extracted from the Cityscapes dataset's annotations. We then placed the patch in the middle of this target area to effectively quantize the area of effect of the adversarial patch. By this we aim to simulate the effect of real-life patches, where they can be found in the middle of objects instead of overstretching to to multiple objects.
Some sample attacks can be seen on Fig. \ref{fig:alg.illustration}.

As these results demonstrate patch based attacks are feasible in practice in case of segmentation problems. Based on this findings one can ask the question what kind of limitations are there for patch based attacks, can they generate arbitrary, general output maps?

\section{A complexity analysis of patch based attacks}

In this section we will prove that it is impossible to generate arbitrary output images by attacking the inputs with patches.
Obviously patch based attacks can not modify pixels which fall outside of the receptive field of corresponding neurons which limits their effectiveness based on the selected architecture. Here we will demonstrate that if patch-based attacks could generate arbitrary outputs, than the size of these patches has to be limited to a much smaller region than their receptive field.

Our proof follows the following way:
We investigate a number of total different segmentation maps that can be generated in a region and provide an upper-bound for them based on the network architecure. We consider two maps different if the winning class (the largest value after softmax normalisation) at least at one pixel is different. This can only happen if non-linear changes happen in the forward path of the network. 
If one considers an output map in one region of the image containing $W \times H$ pixels where $W$ represents the width, $H$ the height of the region and the network used for semantic segmentation can differentiate between $D$ number of classes one can generate: $D^{WH}$ different output maps where at least one pixel is classified differently.

To allow an attack that can generate all these patterns the network has to contain at least this many distinct linear regions (separated by a non-linear change in network output), since changing the largest value at a selected output pixel can only be implemented by non-linear functions, hence it is a non-monotonic change. To simply put it, the network has to move the output to a different linear region to ensure that it generates a different output map.
Also it is not necessary that every linear region generates a different output map and while many of these could be the same, we will demonstrate that even if each one of them would be different, they still can not cover all the possible output maps in practice.

For this we will calculate the maximal number of linear regions in which the patch is involved and we will demonstrate that it is significantly larger than $D^{WH}$, which shows that in practice it is not possible to generate arbitrary output maps in case of semantic segmentation using patch based attacks.

\subsection{Upper bound of linear regions}

An upper bound for the number of liner regions in a fully connected layer using ReLus as non-linearities was first introduced in the paper of Montufar et al in 2014 \cite{montufar2014number}, which states that the number of linear regions $R_n$ is upper bounded by the following expression:
\begin{equation}
R_{L_{FC}} \leq \sum_{i=0}^{n_0} \binom{n_1}{i}
\end{equation}
where the layer contains $n_0$ number of input and $n_1$ number of output neurons.

In 2020 in the paper \cite{xiong2020number} this theorem was extended to convolutional networks where the authors proved that applying $L$ number of consecutive convolutional layers cannot generate more separated linear regions than:
\begin{equation}
R_{Conv} \leq   \prod_{l=1}^{L} \sum_{ i=0 }^{w_{0} h_{0} c_0} \binom{w_{l} h_{l} c_{l}} {i}
\end{equation}
where in an $L$ layered network $w_k$ and $h_k$ represent the spatial dimensions of the data in the $k$-th layer, meanwhile $c_k$ represents the number of channels in the selected layer (e.g. $w_{0}, h_{0}, c_0$ note the dimensions of the input data).

This formula means that a convolutional layer working on an input data of $25 \times 25$ pixels containing 64 channels can multiply the number of linear regions by $3.29*10^{220}$, meanwhile in case of  128 channels this number is $5.3*10^{269}$.

In case of patch-based attacks the spatial dimensions are determined by the receptive field of the neurons in which the original patch is present, but these are typically small numbers, since the patch has to be small to ensure that it is hardly noticeable on the image by human perception.
This means that a typical convolutional layer (containing 128 or 256 channels) where the size of a patch is around  $25\times25$ pixels multiplies the number of linear regions by less than $10^{300}$. This way five convolutional layers could generate (at most) $10^{1500}$ linear regions.
In case of ten possible output classes this means that if this network could generate all possible output elements in a region, the region can not be larger than $10^{WH}$, where $WH$ has to be smaller than 1500. Otherwise the number of possible output maps would be larger and could not be generated by a network with such complexity.
In this case for example it means that a $25x25$ patch could only generate output maps with the area of 1500 pixels (~38x38 pixels), if they are general and can yield arbitrary outputs.

Table \ref{TAbUniversalSizes} contains the maximal number of linear regions for networks which are typically applied on semantic segmentation problems. These data were calculated for different patch sizes along with the maximal patch size which can be generated with this complexity if we assume that patch generation is universal, namely arbitrary output maps can be created with an appropriate attack. The numbers were done considering the Cityscapes dataset as a case study, where each output pixel can belong to nineteen different classes.

\begin{table}
\centering
\caption{This table displays the maximal number of non-linear regions ($R_N$) for different network architectures (UNET\cite{ronneberger2015u}, FCN8\cite{long2015fully}, MobileNetv3-Large ($MN_{V3}$)\cite{hu2018squeeze} and DeeplabV3 with ResNet18 backbone ($DL_{V3}$ )\cite{chen2017rethinking} and ) and patch sizes ($S_R$) along with maximal number of pixels which could be changed by such a region if generic output maps can be created}\label{TAbUniversalSizes}
\begin{tabular}{|l|c|c|c|c|} 
\hline
 & UNET & FCN8 & $MN_{V3}$ & $DL_{V3}$  \\ 
\hline
$R_N$ (2x2) & $10^{219}$ & $10^{168}$ & $10^{229}$ & $10^{584}$   \\ 
$S_R$ (2x2) & 13x13     & 11x11 & 13x13  & 21x21   \\ 
\hline
$R_N$ (5x5) & $10^{1448}$ & $10^{1203}$ & $10^{1239}$ &$10^{3421}$    \\ 
$S_R$ (5x5) & 33x33   &  30x30 & 31x31   & 51x51      \\ 
\hline
$R_N$ (10x10) & $10^{5034}$ &$10^{4646}$& $10^{3446}$ & $10^{12725}$    \\ 
$S_R$ (10x10) & 62x62  &  60x60  & 51x51    & 99x99 \\ 
\hline
$R_N$ (20x20) & $10^{16842}$ & $10^{17864}$ &$10^{9343}$  &$10^{48151}$   \\ 
$S_R$ (20x20) & 114x114 &   118x118 & 85x85    & 194x194    \\ 
\hline
\end{tabular}
\label{table1}
\end{table}

%!Deelab v3 with resnet 50 is fairly complex to generate large pathces
% \hline
% DEEPLAbV3--ResNEt50   & $10^{2054}$ & $10^{11919}$      & $10^{43563}$   &  $10^{161393}$     \\ 
% UniversalPatchSize   & 37x37      & 89x89    & 171x171 & 330x330   
%RESNET 18-as számok!!!

% In Deeplab:v3 it might deepend on the whole backbone and ime size, but it is usually larger than the whoel image...
% 130x130
% 10^17100

We also have to mention that all the aforementioned calculations contain the upper bound of linear regions. This number can only be achieved if a new non-linearity intersects each earlier linear regions.  It was demonstrated in \cite{serra2018bounding} and \cite{trimmel2021tropex} that the number of linear regions can also be measured or algorithmically approximated using sampling in a trained architecture. In all the investigated cases the number of linear regions were significantly smaller in practice than the upper bound provided by the theorem. (e.g. in case of the investigated seven and eight layered convolutional networks the upper bound was $356180$ and $819115$, meanwhile by sampling methods only $3398$ and $4822$ linear regions were identified in the trained networks).
This means that in practice one can assume that the number of output maps which can be generated by a network is orders of magnitude smaller than the previously introduced upper bounds and this way the maximal sizes in which arbitrary output maps can be generated is significantly smaller. The exact measurement of linear regions is important in practice for trained networks, but the presented upper bounds are more general and depend only on the network architecture and not the training data or the weights of the network.

We have seen in the previous sections that adversarial patches affect larger regions than the number presented in Table \ref{TAbUniversalSizes}. Based on these empirical results we can state that patch based attacks can not be general in practice and they can not produce arbitrary output maps. We also have to mention that this paper only shows that arbitrary output maps can not be generated. This does not mean that in practice a whole object could not be altered, non-existing objects could be hallucinated or one could not make a person disappear in semantic segmentation in practice using patches as camouflage, which can also be seen on some of our samples. 

\section{Conclusion}
 In this paper we have investigated the generality of patch based attacks in semantic segmentation problems. For our case study we have investigated a simple simulated dataset with the U-NET architecture and the commonly investigated Cityscapes datasets and two commonly used network architectures: DeepLab V3 and MobileNet V3. Our finding shows that patch based attacks are feasible in case of semantic segmentation in practice and even in case of small patches such as 2x2 they are able to change the output classes of the segmentation maps on a large area (in certain cases containing more than 5000 modified pixels).
 On the other hand the complexity analysis of these architectures revealed that these networks can generate a lower number of possible outputs maps.
 This deducts that semantic segmentation can only be attacked successfully under certain conditions with patch based attacks, which largely depend on the network architecture and patch based attacks can not results arbitrary output maps. There are certain output maps in these segmentation networks which can not be generated by any patches (with limited size) for a given input.
 
\vspace{-15pt}
\section*{Acknowledgment}
\vspace{-10pt}
 This research has been partially supported by the Hungarian Government by the following grant: 2018-1.2.1-NKP00008: Exploring the Mathematical Foundations of Artificial Intelligence and the support of the Alfréd Rényi Institute of Mathematics if also gratefully acknowledged.
 \vspace{-15pt}
\bibliographystyle{ieeetr}
\bibliography{adv}

\end{document}